\documentclass{article}
\usepackage{amsmath}
\usepackage{arxiv}
\usepackage[utf8]{inputenc} 
\usepackage[T1]{fontenc}    
\usepackage{hyperref}       
\usepackage{url}            
\usepackage{booktabs}       
\usepackage{amsfonts}       
\usepackage{nicefrac}       
\usepackage{microtype}      
\usepackage{graphicx}
\graphicspath{ {./images/} }
\usepackage{graphicx}  
\usepackage{placeins}  

\title{PL-FGSA: A Prompt Learning Framework for Fine-Grained Sentiment Analysis Based on MindSpore}

\author{
  Zhenkai Qin$^{1,2,3}$ \thanks{These authors contributed equally to this work.} \\
  $^{1}$School of Information Technology \\
  $^{2}$Network Security Research Center \\
  $^{3}$Big Data and Policing Technology Laboratory\\
  Guangxi Police College\\
  Nanning, Guangxi, China \\
  \texttt{qinzhenkai@gxjcxy.edu.cn} \\
  \And
  Jiajing He$^{2}$ \thanks{These authors contributed equally to this work.}\\
  School of Information Technology \\
  Guangxi Police College \\
  Nanning, Guangxi, China \\
  \texttt{hejiajing@gxjcxy.edu.cn} \\
  \And
  Qiao Fang$^{2}$ \thanks{These authors contributed equally to this work.} \\
  School of Information Technology \\
  Guangxi Police College \\
  Nanning, Guangxi, China \\
  \texttt{fangqiao@gxjcxy.edu.cn} \\
}

\begin{document}
\maketitle

\begin{abstract}
Fine-grained sentiment analysis (FGSA) aims to identify sentiment polarity toward specific aspects within a text, enabling more precise opinion mining in domains such as product reviews and social media. However, traditional FGSA approaches often require task-specific architectures and extensive annotated data, limiting their generalization and scalability. To address these challenges, we propose PL-FGSA, a unified prompt learning-based framework implemented using the MindSpore platform, which integrates prompt design with a lightweight TextCNN backbone. Our method reformulates FGSA as a multi-task prompt-augmented generation problem, jointly tackling aspect extraction, sentiment classification, and causal explanation in a unified paradigm. By leveraging prompt-based guidance, PL-FGSA enhances interpretability and achieves strong performance under both full-data and low-resource conditions. Experiments on three benchmark datasets—SST-2, SemEval-2014 Task 4, and MAMS—demonstrate that our model consistently outperforms traditional fine-tuning methods and achieves F1-scores of 0.922, 0.694, and 0.597, respectively. These results validate the effectiveness of prompt-based generalization and highlight the practical value of PL-FGSA for real-world sentiment analysis tasks.
\end{abstract}


\section{Introduction}

In recent years, fine-grained sentiment analysis (FGSA) has emerged as a critical task in natural language processing (NLP), aiming to detect sentiment polarities associated with specific aspects mentioned in a given text. Unlike coarse-grained sentiment classification, which considers the entire text as a single unit, FGSA captures the nuanced opinions embedded in user reviews, product descriptions, and social media content~\cite{tang2019aspect}. This task plays a vital role in real-world applications such as customer feedback analysis, brand monitoring, and intelligent recommendation systems.

Traditional approaches to FGSA often decompose the problem into several subtasks, including aspect term extraction (ATE)~\cite{augustyniak2021comprehensive}, aspect sentiment classification (ASC)~\cite{brauwers2022survey}, and, more recently, causal explanation generation (CEG)~\cite{zhang2023ceg}. Most of these methods rely on fully supervised learning and task-specific architectures such as BiLSTM, attention mechanisms, or dependency parsers. While effective to some extent, these models face two major limitations: (1) a strong dependence on large-scale annotated datasets, which are expensive and labor-intensive to obtain; and (2) limited generalization across domains, especially under low-resource or few-shot learning conditions.

Prompt learning has recently gained significant attention as a lightweight yet powerful paradigm that leverages pre-trained language models (PLMs) by reformulating downstream tasks into masked or cloze-style text generation problems~\cite{li2022prompt}. Unlike traditional fine-tuning, which modifies a large number of parameters, prompt learning can perform well even with minimal task-specific tuning. However, the application of prompt learning to FGSA—particularly to multi-task settings involving both structured prediction (e.g., aspect extraction) and natural language generation (e.g., causal explanation)—remains underexplored.

To address these challenges, we propose PL-FGSA, a novel prompt learning-based framework for fine-grained sentiment analysis. Implemented using the MindSpore platform, PL-FGSA integrates a prompt-enhanced TextCNN encoder to support joint modeling of three critical FGSA subtasks: aspect extraction, sentiment classification, and causal explanation. Instead of designing separate modules for each task, we unify them under a multi-task prompt formulation. Specifically, task-specific prompts are designed to guide the model in identifying aspect terms, inferring sentiment polarities, and generating natural language explanations. This unified structure improves both interpretability and parameter efficiency.

We evaluate PL-FGSA on three benchmark datasets—SST-2, SemEval-2014 Task 4, and MAMS—covering both sentence-level and aspect-level sentiment tasks. Experimental results show that our model consistently outperforms traditional fine-tuning methods, achieving F1-scores of 0.922, 0.694, and 0.597, respectively. Furthermore, PL-FGSA demonstrates strong generalization under few-shot settings and provides transparent, human-readable rationales for its predictions.

The main contributions of this paper are summarized as follows:
\begin{itemize}
    \item We propose PL-FGSA, a unified framework for fine-grained sentiment analysis that integrates prompt learning with a lightweight TextCNN backbone to jointly perform aspect extraction, sentiment classification, and causal explanation.
    \item We design a prompt-enhanced multi-task modeling strategy that supports both structured prediction and natural language generation in a unified architecture.
    \item Extensive experiments on three FGSA benchmarks validate the effectiveness and robustness of PL-FGSA under both full-data and few-shot scenarios.
\end{itemize}

\section{Related Work}

\subsection{Fine-Grained Sentiment Analysis}
Fine-grained sentiment analysis (FGSA) is a specialized task in sentiment analysis that seeks to determine the polarity of opinions with respect to specific aspects, targets, or entities mentioned in text. Traditional approaches treated FGSA as a pipeline process, where aspect extraction and sentiment classification were handled separately. Pontiki et al.~\cite{pontiki2016semeval} formalized the benchmark setup for aspect-based sentiment analysis and provided early baselines based on statistical learning. Subsequently, Huang et al.~\cite{huang2020syntax} introduced a syntax-aware graph attention network that leverages syntactic structure to enhance the interaction between aspect terms and their contextual semantics. Li et al.~\cite{li2019unified} proposed a unified model capable of jointly extracting opinion targets and predicting their sentiment polarities, demonstrating the benefits of end-to-end modeling in fine-grained sentiment tasks. Although these methods show strong performance in supervised settings, they suffer from limitations in scalability, domain transfer, and generalization when large annotated datasets are unavailable. These challenges motivate the exploration of lightweight and data-efficient alternatives such as prompt learning.

\subsection{Prompt Learning in Sentiment Analysis}
Prompt learning has emerged as an effective paradigm for adapting pre-trained language models (PLMs) to downstream NLP tasks, especially in few-shot and zero-shot scenarios. Instead of fine-tuning the entire model, prompt-based methods reformulate the task into a masked language modeling objective, allowing the model to leverage its pre-trained knowledge. Mao et al.~\cite{mao2022biases} conducted an extensive empirical study revealing that prompt-based sentiment and emotion classification methods are highly sensitive to prompt formulation and exhibit systematic biases inherited from PLMs. Liu et al.~\cite{liu2021p} advanced prompt learning with P-Tuning, which replaces discrete templates with continuous embeddings, achieving competitive performance with minimal supervision. Sun et al.~\cite{sun2024harnessing} further proposed a prompt knowledge tuning method that integrates domain-specific insights into aspect-based sentiment analysis, significantly improving the model's interpretability and generalization in real-world tasks. Despite these advances, most prompt learning methods are applied to sentence-level sentiment classification, and there remains limited exploration in aspect-aware or fine-grained sentiment contexts. Moreover, the current ecosystem is tightly coupled with PyTorch and HuggingFace, making it challenging to deploy prompt-based models efficiently in diverse hardware environments.

\subsection{Causal and Explanation-Enhanced Sentiment Models}
The increasing demand for transparent and interpretable NLP systems has led to the integration of causal reasoning and explanation mechanisms into sentiment models. Zhang et al.~\cite{zhang2019aspect} developed LCF-BERT, which distinguishes between local and global context around each aspect term, enhancing both classification accuracy and explainability. Zhou et al.~\cite{zhou2023causalabsc} proposed CausalABSC, a causality-aware framework for aspect-based sentiment classification that leverages causal inference to mitigate aspect-level sentiment bias. These models highlight the benefits of integrating reasoning pathways into deep networks, particularly for decision-critical domains such as healthcare and finance. However, they often require additional annotated resources for explanations or causal labels, and their complexity limits scalability in real-time applications. Although post-hoc explanation techniques like SHAP and LIME are sometimes used, they tend to lack the fidelity and task-specific alignment required in FGSA tasks. This presents an ongoing challenge in designing efficient and inherently interpretable models.

\subsection{MindSpore-Based Sentiment Modeling}
MindSpore is an open-source deep learning framework developed by Huawei, designed for efficient training and deployment across cloud, edge, and device environments. Its support for static computational graphs, operator fusion, and memory-efficient scheduling makes it particularly suitable for lightweight and modular AI applications. According to Huawei Technologies~\cite{tong2021study}, MindSpore introduces a Cell-based abstraction system that simplifies model encapsulation and reuse, offering architectural flexibility not easily achievable in other frameworks. While PyTorch and TensorFlow dominate sentiment analysis research, MindSpore's integration with Ascend AI chips and mobile devices provides unique deployment advantages. Qin et al.~\cite{qin2025few} proposed a few-shot hate speech detection framework built on MindSpore, demonstrating high efficiency and strong adaptability under low-resource conditions. However, current studies have not addressed the challenges of integrating prompt learning with aspect-level sentiment tasks under this framework. This lack of research limits the adoption of prompt-based FGSA models in real-world production pipelines, especially in scenarios requiring national hardware compatibility or restricted data environments.

\subsection{Our Motivation: Prompt Learning under the MindSpore Framework}
In light of the limitations described above, this work introduces PL-FGSA: a Prompt Learning framework for Fine-Grained Sentiment Analysis implemented natively in MindSpore. Unlike existing methods that rely on heavy PyTorch-based pipelines, PL-FGSA leverages MindSpore's graph-optimized execution and modular design to build a fully integrated architecture consisting of prompt-aware encoders, aspect-conditioned sentiment predictors, and explanation generators. This framework aims to (i) reduce reliance on large-scale labeled datasets through few-shot prompting, (ii) enhance model interpretability by coupling sentiment decisions with causal explanations, and (iii) enable deployment across resource-constrained platforms including CPU-only systems and Ascend NPUs. To the best of our knowledge, PL-FGSA is the first to systematically explore prompt-based fine-grained sentiment modeling in the MindSpore ecosystem, offering a new perspective on scalable, interpretable, and industrial-grade sentiment analysis.

\section{Method}

In this section, we describe the architecture and learning strategy of PL-FGSA, a unified and prompt-conditioned framework for Fine-Grained Sentiment Analysis (FGSA). The model is designed to simultaneously address three interrelated subtasks—Aspect Term Extraction (ATE), Aspect Sentiment Classification (ASC), and Causal Explanation Generation (CEG)—through a shared input representation and a multi-task learning mechanism.

The overall architecture of PL-FGSA comprises three main components: (1) a \textbf{Prompt-Conditioned Input Construction} module that injects task-specific instructions into natural language sequences, (2) a \textbf{TextCNN-Based Feature Encoder} that captures local semantic dependencies using multi-scale convolutional filters, and (3) \textbf{Unified Multi-Task Prediction Heads} that specialize in handling the outputs for each subtask. All modules are implemented in the MindSpore deep learning framework, enabling scalable, efficient training and deployment across diverse computing environments.

\begin{figure}[htbp]
    \centering
    \includegraphics[width=0.95\linewidth]{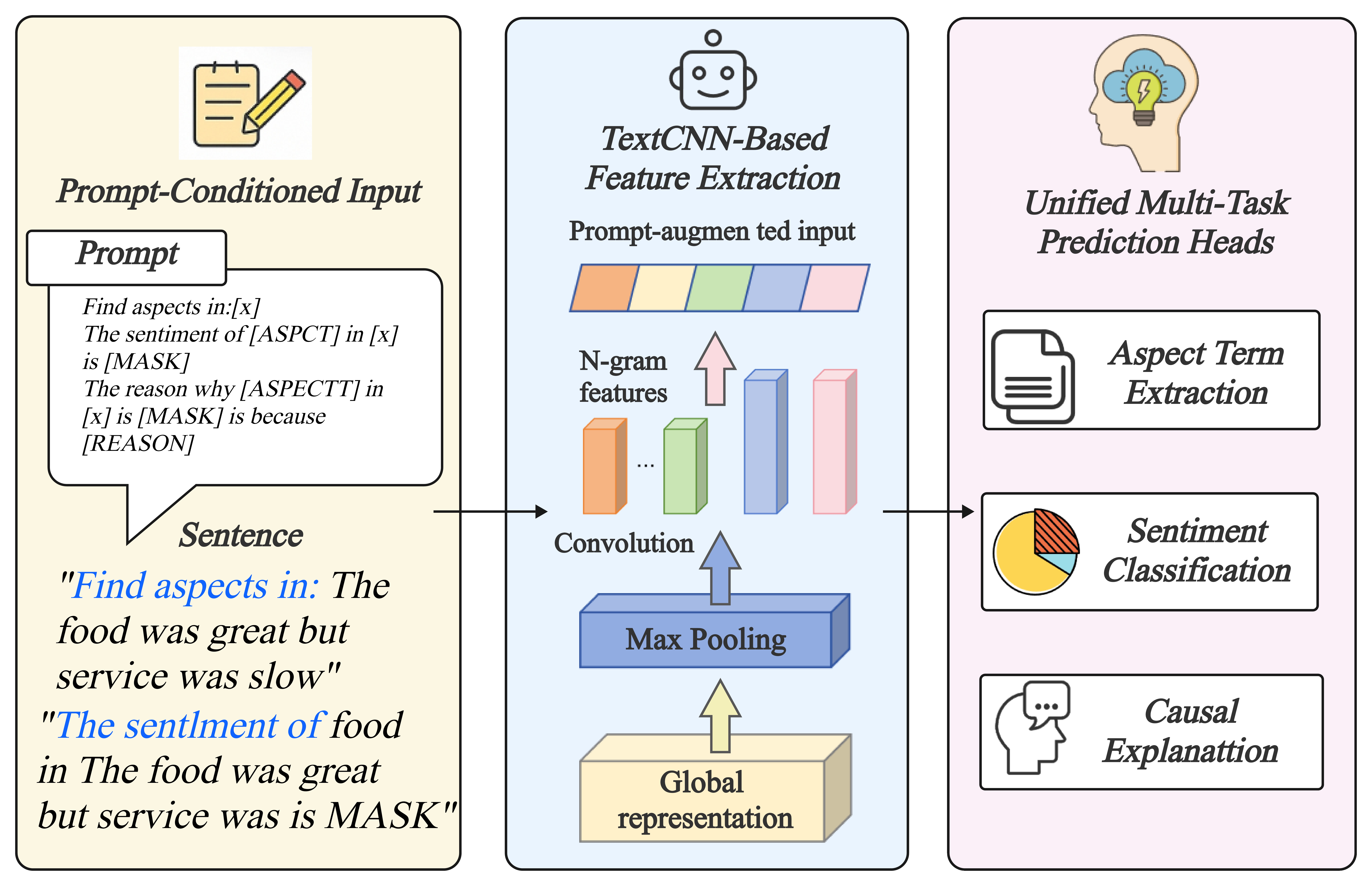}
    \caption{The overall architecture of the PL-FGSA framework. Task-specific prompts are prepended to raw sentences to guide the model across different subtasks. The shared encoder extracts local features via convolution and max pooling, which are then passed to dedicated output heads for ATE, ASC, and CEG.}
    \label{fig:model-architecture}
\end{figure}

Figure~\ref{fig:model-architecture} illustrates the overall pipeline of PL-FGSA. Given a raw sentence, task-specific prompts are first constructed and embedded alongside the input tokens to form a prompt-augmented sequence. This sequence is then passed through a lightweight convolutional encoder to extract n-gram features and obtain a compact global representation. Finally, this representation is routed to three parallel output heads responsible for sequence labeling (ATE), classification (ASC), and generation (CEG). The modularity of the architecture allows efficient parameter sharing, promotes cross-task synergy, and supports flexible extension to additional subtasks if needed.

\subsection{Prompt-Conditioned Input Construction}

Traditional FGSA models often rely on task-specific input formats and classifiers, resulting in fragmented architectures and limited generalization. To overcome this, PL-FGSA employs a prompt learning strategy to unify input representations across tasks by encoding task intent directly into the input sequence.

Given an input sentence $x = \{w_1, w_2, ..., w_n\}$, we construct task-specific prompt templates to reformulate the original input into a textual instruction format:

\begin{itemize}
    \item \textbf{ATE:} ``\textit{Find aspects in: [x]}'' 
    \item \textbf{ASC:} ``\textit{The sentiment of [ASPECT] in [x] is [MASK]}'' 
    \item \textbf{CEG:} ``\textit{The reason why [ASPECT] in [x] is [MASK] is because [REASON]}'' 
\end{itemize}

These prompts embed explicit task semantics into the input space, eliminating the need for separate architectures and enabling effective instruction-following behavior within a single model.

\begin{figure}[htbp]
    \centering
    \includegraphics[width=0.9\linewidth]{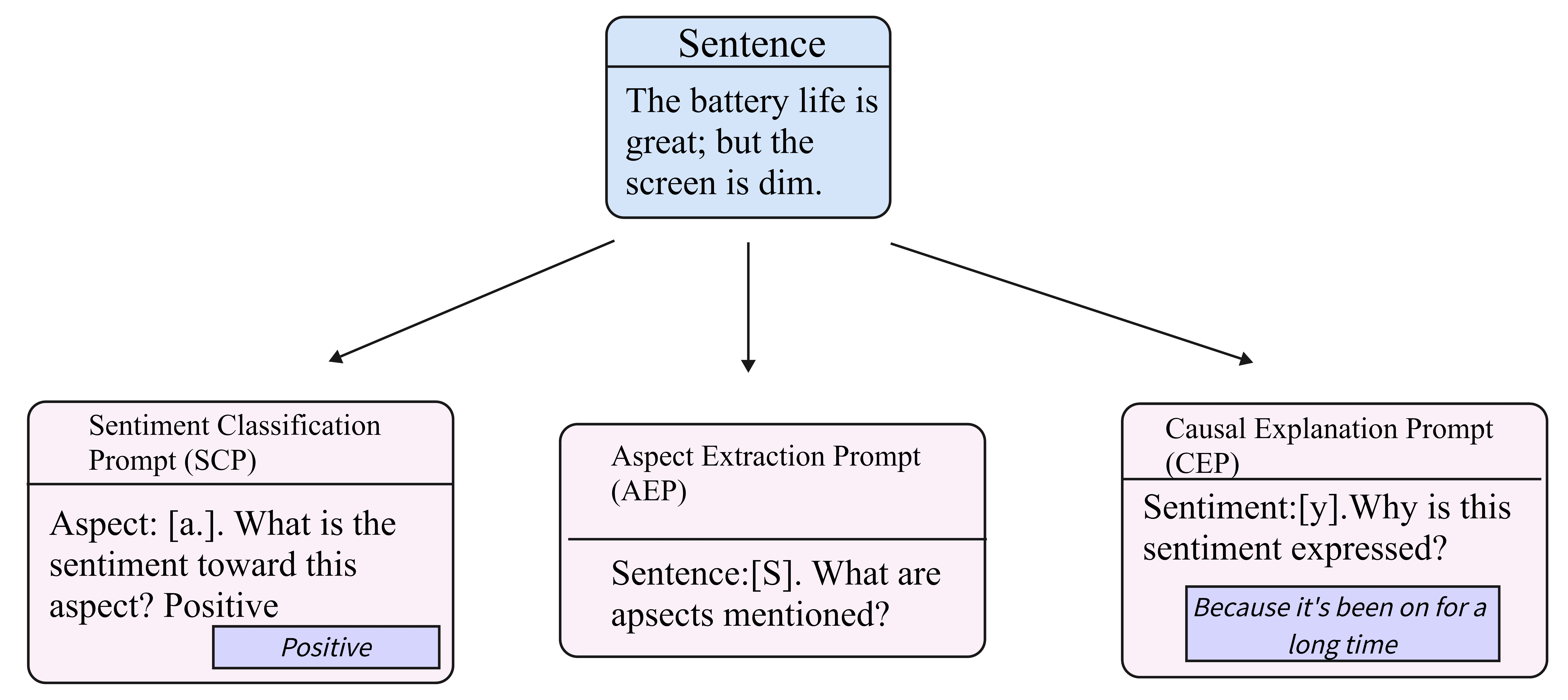}
    \caption{An illustrative example of prompt-conditioned input construction for three FGSA subtasks: aspect extraction (AEP), sentiment classification (SCP), and causal explanation generation (CEP).}
    \label{fig:prompt-example}
\end{figure}

To better illustrate how the prompts guide different subtask formulations, an example is shown in Figure~\ref{fig:prompt-example}. Given the sentence “The battery life is great; but the screen is dim,” the model applies distinct prompts for aspect extraction, sentiment classification, and causal explanation generation. Each prompt rephrases the original sentence to reflect the specific reasoning objective while maintaining a consistent input structure.

\subsection{TextCNN-Based Feature Encoder}

To capture local semantic dependencies within prompt-enhanced inputs, we utilize a TextCNN module as the backbone encoder. Compared to recurrent architectures, CNNs offer high parallelism and lower inference latency, which is beneficial for deployment in real-time sentiment analysis applications.

The encoder applies 1D convolutional filters with kernel sizes $k \in \{3,4,5\}$ across the input matrix $\mathbf{X}$ to extract n-gram features of varying lengths:

\begin{equation}
    \mathbf{c}_k = \text{ReLU}(\mathbf{X} * \mathbf{W}_k + \mathbf{b}_k),
\end{equation}

where $\mathbf{W}_k \in \mathbb{R}^{k \times d}$ is the kernel matrix, and $\mathbf{b}_k$ is a learnable bias term. Each resulting feature map $\mathbf{c}_k$ undergoes max-over-time pooling:

\begin{equation}
    \mathbf{h}_k = \max_{i} \mathbf{c}_k[i],
\end{equation}

and all pooled features are concatenated to form a global representation:

\begin{equation}
    \mathbf{h} = [\mathbf{h}_3; \mathbf{h}_4; \mathbf{h}_5] \in \mathbb{R}^{3d}.
\end{equation}

This vector $\mathbf{h}$ serves as the high-level semantic representation for classification and generation tasks.

\subsection{Multi-Task Output Heads}

To accommodate the diverse output formats of the three FGSA subtasks, PL-FGSA adopts a branched architecture that shares a common encoder but uses task-specific heads. This design enables efficient parameter sharing while preserving task specialization.

\textbf{(1) Aspect Term Extraction (ATE):}  
ATE is modeled as a sequence labeling task. A linear layer followed by softmax is applied to each token embedding in $\mathbf{X}$ to predict BIO tags:

\begin{equation}
    \hat{Y}^{\text{ATE}} = \text{softmax}(\mathbf{X} \cdot \mathbf{W}^{\text{ATE}} + \mathbf{b}^{\text{ATE}}),
\end{equation}

where $\mathbf{W}^{\text{ATE}} \in \mathbb{R}^{d \times L}$ and $L$ is the number of label classes.

\textbf{(2) Aspect Sentiment Classification (ASC):}  
For ASC, the global representation $\mathbf{h}$ is used to predict the sentiment polarity of a given aspect through a fully connected layer:

\begin{equation}
    \hat{y}^{\text{ASC}} = \text{softmax}(\mathbf{W}^{\text{ASC}} \cdot \mathbf{h} + \mathbf{b}^{\text{ASC}}),
\end{equation}

where $\mathbf{W}^{\text{ASC}} \in \mathbb{R}^{3d \times C}$ and $C$ is the number of sentiment categories.

\textbf{(3) Causal Explanation Generation (CEG):}  
CEG aims to generate natural language explanations for a given sentiment decision. A shallow GRU-based decoder $D_{\text{CEG}}$ is initialized with $\mathbf{h}$ and produces output tokens autoregressively:

\begin{equation}
    P(y_{1:T} | \mathbf{h}) = \prod_{t=1}^{T} P(y_t | y_{<t}, \mathbf{h}),
\end{equation}

This decoder allows the model to produce free-form justifications for its predictions, improving interpretability.

\subsection{Joint Training Objective}

Multi-task learning is employed to jointly optimize the three output heads, leveraging the shared encoder and maximizing cross-task information synergy. The total loss function is defined as:

\begin{equation}
    \mathcal{L}_{\text{total}} = \lambda_1 \mathcal{L}_{\text{ATE}} + \lambda_2 \mathcal{L}_{\text{ASC}} + \lambda_3 \mathcal{L}_{\text{CEG}},
\end{equation}

where $\lambda_i$ are manually assigned or learned weights for balancing task contributions. For ATE and ASC, we use standard categorical cross-entropy, while CEG is trained using sequence-level negative log-likelihood:

\begin{equation}
    \mathcal{L}_{\text{CEG}} = -\sum_{t=1}^{T} \log P(y_t | y_{<t}, \mathbf{h}).
\end{equation}

This formulation encourages joint feature learning, reduces task interference, and improves generalization under data scarcity.

\subsection{Implementation in MindSpore}

To ensure reproducibility, scalability, and deployment readiness, PL-FGSA is implemented using MindSpore 2.5.0. All components, including the prompt processor, encoder, and task heads, are implemented as modular \texttt{nn.Cell} units.

The training process adopts the Adam optimizer with an initial learning rate of $1 \times 10^{-3}$ and cosine annealing for dynamic adjustment. Mixed-precision training (AMP) and graph fusion are enabled to enhance computational efficiency. Batch size is set to 16. The model supports export to ONNX format and deployment on Ascend hardware, enabling low-latency, high-throughput inference in practical industrial settings.

\section{Experimental Results and Analysis}

\subsection{Datasets}

To evaluate the generalization capability of PL-FGSA across diverse sentiment analysis tasks, we select three benchmark datasets with varying levels of semantic granularity:

\textbf{SST-2}: The Stanford Sentiment Treebank 2 is a binary sentiment classification dataset where each sentence is labeled as either positive or negative. It represents the canonical sentence-level sentiment analysis task with well-balanced labels and clearly expressed sentiment cues.

\textbf{SemEval-2014 Task 4}: This dataset consists of aspect-level sentiment annotations for restaurant and laptop reviews. Each instance may contain multiple aspect terms with corresponding sentiment polarities. The task challenges include implicit aspect references, overlapping spans, and aspect-sentiment contextual ambiguity~\cite{kirange2014aspect}.

\textbf{MAMS}: The Multi-Aspect Multi-Sentiment dataset extends the complexity by including reviews that contain multiple aspect terms with potentially contrasting sentiment labels within a single sentence. It is designed to evaluate a model's ability to reason about fine-grained and contrastive opinion expressions, making it one of the most challenging FGSA datasets.

All datasets are preprocessed using standardized tokenization. Official train/dev/test splits are retained for fair comparison, and prompt templates are constructed manually for each task type to ensure consistent semantic framing.

\subsection{Experimental Settings}

Experiments are conducted on a CPU-only environment using MindSpore 2.5.0 and Python 3.9. The hardware configuration includes an Intel Xeon processor and 64 GB of RAM. All models are trained for 10 epochs using the Adam optimizer with an initial learning rate of 0.01 and a batch size of 16.

\subsection{Evaluation Metrics}

We report four widely used metrics to evaluate the model: Accuracy, Precision, Recall, and F1-score. Each metric provides a different perspective on model behavior:

\textbf{Accuracy (Acc)} measures the overall rate of correct predictions, offering a general sense of performance across all classes:
\begin{equation}
\text{Accuracy} = \frac{TP + TN}{TP + TN + FP + FN},
\end{equation}

\textbf{Precision (Prec)} evaluates the correctness of predicted positive instances:
\begin{equation}
\text{Precision} = \frac{TP}{TP + FP}.
\end{equation}

\textbf{Recall (Rec)} quantifies the model's ability to retrieve all relevant instances:
\begin{equation}
\text{Recall} = \frac{TP}{TP + FN}.
\end{equation}

\textbf{F1-score (F1)} harmonizes precision and recall into a single score:
\begin{equation}
\text{F1-score} = 2 \times \frac{\text{Precision} \times \text{Recall}}{\text{Precision} + \text{Recall}}.
\end{equation}

All metrics are reported as macro-averages across classes. In particular, the F1-score is emphasized in multi-aspect tasks such as SemEval and MAMS, where label imbalance and aspect polarity ambiguity are prevalent.

\subsection{Results and Analysis}

\begin{figure}[htbp]
    \centering
    \includegraphics[width=0.9\linewidth]{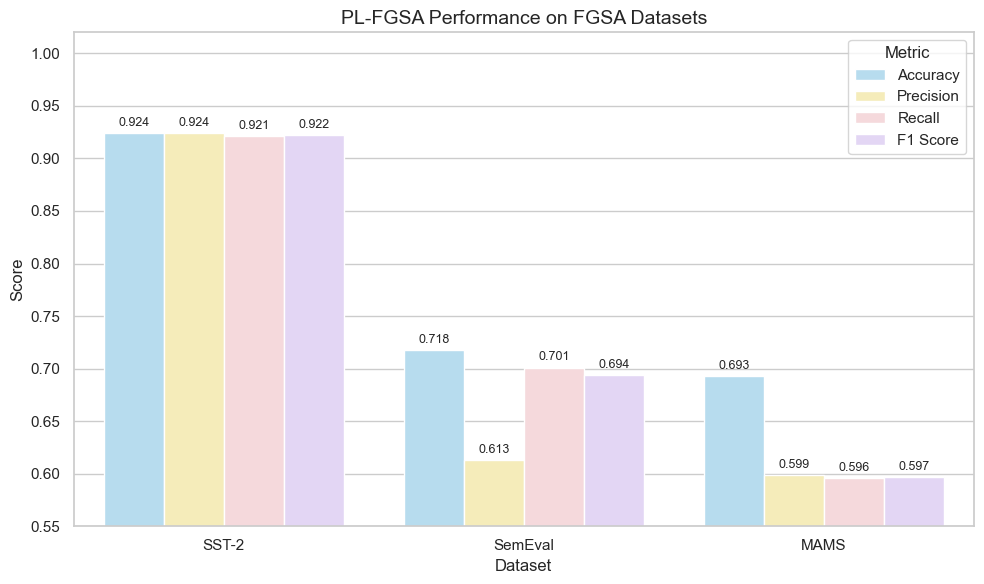}
    \caption{Visualization of PL-FGSA’s performance on SST-2, SemEval, and MAMS datasets across four evaluation metrics: Accuracy, Precision, Recall, and F1 Score.}
    \label{fig:results-bar}
\end{figure}

The quantitative performance of PL-FGSA across three benchmark datasets—SST-2, SemEval-2014 Task 4, and MAMS—is summarized in Table~\ref{tab:results}. We report four commonly used evaluation metrics: Accuracy, Precision, Recall, and F1 Score, all computed under the macro-averaging scheme to account for class imbalance,The comparative results are further visualized in Figure~\ref{fig:results-bar}.

\begin{table}[htbp]
\centering
\caption{Performance of PL-FGSA on SST-2, SemEval-2014, and MAMS datasets.}
\label{tab:results}
\begin{tabular}{@{}lcccc@{}}
\toprule
\textbf{Dataset} & \textbf{Accuracy} & \textbf{Precision} & \textbf{Recall} & \textbf{F1 Score} \\
\midrule
SST-2   & 0.924 & 0.924 & 0.921 & 0.922 \\
SemEval & 0.718 & 0.613 & 0.701 & 0.694 \\
MAMS    & 0.693 & 0.599 & 0.596 & 0.597 \\
\bottomrule
\end{tabular}
\end{table}

As shown, PL-FGSA attains strong performance on the SST-2 dataset, achieving over 0.924 on all four metrics. This confirms the model's effectiveness in traditional sentence-level sentiment classification tasks, where sentiment polarity is often explicit, and linguistic structure is relatively simple. The integration of prompt-conditioned inputs and lightweight convolutional encoding enables efficient and robust feature extraction, particularly beneficial for direct sentiment scenarios like SST-2.

For the SemEval-2014 and MAMS datasets, which focus on fine-grained aspect-level sentiment analysis, PL-FGSA demonstrates consistent and competitive performance. The macro-F1 scores of 0.694 and 0.597 highlight the framework's capacity to manage more complex linguistic phenomena such as aspect co-occurrence, sentiment contrast between different opinion targets, and implicit reasoning over context. These datasets are more challenging due to the presence of multiple overlapping aspects and nuanced emotional cues, making the observed results noteworthy.

Moreover, an interesting observation is that on both SemEval and MAMS, the model yields higher recall than precision. This suggests that PL-FGSA is more inclined to maximize the detection of all potential sentiment-bearing components, even if it occasionally incurs false positives. Such behavior is preferable in practical applications like opinion mining, product review analysis, or customer feedback extraction, where the cost of missing relevant sentiment expressions often outweighs the cost of over-identification.

Overall, the results validate the generalization capability of PL-FGSA across both coarse-grained and fine-grained sentiment tasks. The unified architecture, empowered by prompt learning and multi-task training, proves effective in handling diverse input formats and decision requirements within a single framework.

\section{Discussion}
The experimental results demonstrate that PL-FGSA consistently delivers strong performance across sentence-level, aspect-level, and multi-aspect sentiment classification tasks. This effectiveness can be attributed not only to the integration of prompt learning into a unified architecture but also to the computational efficiency afforded by the MindSpore framework. Through features such as static graph compilation and operator-level optimization, MindSpore significantly reduces training overhead and supports smooth execution even under CPU-only conditions. The modular Cell-based API further facilitates the seamless integration and deployment of PL-FGSA in various practical environments, ensuring scalability and adaptability.

Compared to conventional fine-tuning approaches or task-specific models, PL-FGSA exhibits several distinct advantages. First, the prompt-conditioned design enables a unified modeling paradigm for multiple subtasks—including aspect extraction, sentiment classification, and causal explanation—within a single framework. This avoids the need for isolated modules and enhances inter-task synergy. Second, the use of a lightweight TextCNN backbone, combined with handcrafted prompt templates, ensures a low-complexity yet interpretable representation learning pipeline, particularly suited for low-resource and high-efficiency scenarios. Third, the incorporation of causal explanation generation provides human-understandable rationales for sentiment decisions, significantly improving the interpretability and transparency of the model.

Despite these merits, several limitations remain. The current experimental setup is confined to single-device CPU environments, which may not fully reveal the framework's performance under large-scale distributed scenarios. Additionally, while PL-FGSA achieves competitive results on three well-known benchmarks, its generalization ability across multilingual and domain-shifted datasets warrants further exploration. Moreover, the prompt templates used in this study are manually crafted and static; introducing learnable or context-aware prompt generation strategies could further enhance task adaptability and downstream performance. These directions represent valuable opportunities for future improvement.

The integration of PL-FGSA with the MindSpore ecosystem also highlights the practical potential of combining prompt learning with efficient, industrial-grade AI platforms. MindSpore’s support for mixed precision, automatic optimization, and flexible deployment makes it well-suited for scaling prompt-based models in production environments. Overall, this synergy not only validates the effectiveness of PL-FGSA as a unified sentiment analysis framework but also paves the way for its application in real-world NLP systems where interpretability, generalization, and deployment efficiency are key concerns.

\section{Conclusion}

In this study, we proposed PL-FGSA, a prompt learning framework for fine-grained sentiment analysis that unifies sentence-level, aspect-level, and multi-aspect sentiment classification through a task-agnostic design. Built upon the MindSpore framework, PL-FGSA leverages prompt-based modeling to reduce reliance on large-scale annotations while maintaining adaptability across diverse sentiment structures.

Extensive experiments on SST-2, SemEval-2014 Task 4, and MAMS datasets demonstrate that PL-FGSA achieves consistently strong performance under low-resource and CPU-only settings. The framework exhibits robust generalization across different sentiment granularities without requiring task-specific architectural changes or intensive parameter tuning.

The contributions of this work are three-fold: (i) introducing a unified prompt learning framework that supports multi-level sentiment reasoning without relying on fine-tuned layers; (ii) achieving high interpretability and flexibility across heterogeneous tasks through template-based input design; and (iii) validating the effectiveness and efficiency of PL-FGSA under the MindSpore framework, making it suitable for real-world deployment in constrained environments.

Future research will focus on exploring dynamic or learnable prompt generation strategies, extending PL-FGSA to cross-domain and multilingual sentiment analysis, and integrating soft-prompt tuning to further improve generalization and scalability in complex real-world scenarios.

\section*{Acknowledgments}
Thanks for the support provided by the MindSpore Community.

\vspace{6pt}

\bibliographystyle{unsrt}
\bibliography{references}

\end{document}